\documentclass[10pt,twocolumn,letterpaper]{article}

\usepackage{wacv}
\usepackage{times}
\usepackage{epsfig}
\usepackage{graphicx}
\usepackage{amsmath}
\usepackage{amssymb}
\usepackage{booktabs}
\usepackage{hyperref}
\hypersetup{
	colorlinks = true, 
	urlcolor = magenta, 
	linkcolor = red, 
	citecolor = blue}

\usepackage{enumitem}
\usepackage{romannum}
\usepackage{booktabs}
\usepackage{amsmath}
\usepackage{cuted}
\usepackage{capt-of}

\usepackage{array}
\newcolumntype{P}[1]{>{\centering\arraybackslash}p{#1}}

\usepackage{xcolor,colortbl}
\definecolor{Gray}{gray}{0.9}
\usepackage{cite}

\def\wacvPaperID{1334} 

\wacvalgorithmstrack   

\wacvfinalcopy 

\begin{document}

\title{Pixel-Level Equalized Matching for Video Object Segmentation}
\author{Suhwan Cho$^1$\quad Woo Jin Kim$^1$\quad MyeongAh Cho$^1$\quad Seunghoon Lee$^1$\\
Minhyeok Lee$^1$\quad Chaewon Park$^1$\quad Sangyoun Lee$^{1,2}$\vspace{0.5cm}\\
    $^1$~~Yonsei University\\
    $^2$~~Korea Institute of Science and Technology (KIST)}
\maketitle

\pagenumbering{gobble} 

\begin{strip}
    \vspace{-1.5cm}
    \centering
    \includegraphics[width=1.0\textwidth]{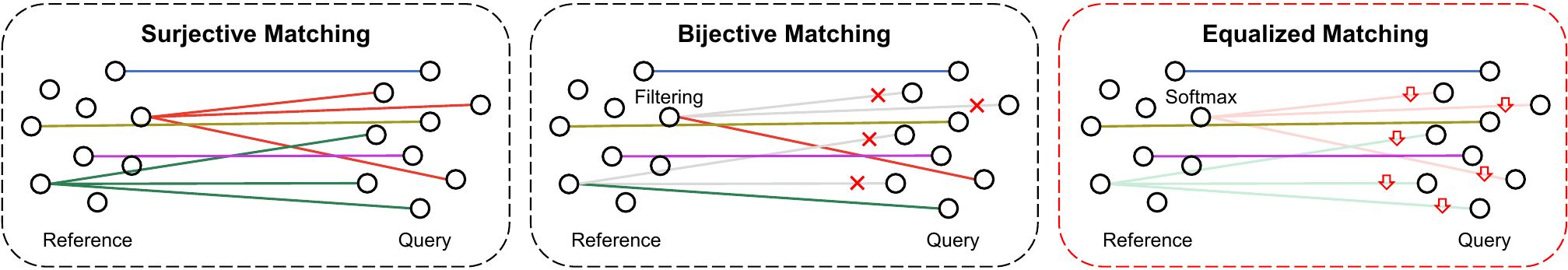}
    \captionof{figure}{Visualized comparison of various feature matching mechanisms. When reference frame information is overly transferred, bijective matching discards uncertain matches to obtain bijectiveness. By contrast, equalized matching can deal with this situation naturally, as all reference frame information is forced to contribute to the query frame equally.\label{figure1}}
\end{strip}

\begin{abstract}
Feature similarity matching, which transfers the information of the reference frame to the query frame, is a key component in semi-supervised video object segmentation. If surjective matching is adopted, background distractors can easily occur and degrade the performance. Bijective matching mechanisms try to prevent this by restricting the amount of information being transferred to the query frame, but have two limitations: 1) surjective matching cannot be fully leveraged as it is transformed to bijective matching at test time; and 2) test-time manual tuning is required for searching the optimal hyper-parameters. To overcome these limitations while ensuring reliable information transfer, we introduce an equalized matching mechanism. To prevent the reference frame information from being overly referenced, the potential contribution to the query frame is equalized by simply applying a softmax operation along with the query. On public benchmark datasets, our proposed approach achieves a comparable performance to state-of-the-art methods.
\end{abstract}

\section{Introduction}
Semi-supervised video object segmentation (VOS) is one of the most important and fundamental tasks in understanding videos. Owing to their strong abilities in tracking and segmenting designated objects, VOS models are widely used in many real-world applications, such as autonomous driving, robotics, sports analytics, and video editing.

In semi-supervised VOS, information about the target objects, i.e., segmentation masks, is only given at the frames where those objects appear for the first time. By utilizing the given information, all designated objects should be segmented over the subsequent frames of a video. A very straightforward approach to achieve this is to find the best-matching pixels of the unannotated pixels (query frame pixels) among the annotated pixels (reference frame pixels), such as in FEELVOS~\cite{FEELVOS} and CFBI~\cite{CFBI}. Once the optimal matches from the query frame to the reference frame are obtained, a query frame segmentation mask can be predicted by transferring reference frame information to the query frame. This matching mechanism is called surjective matching, as only the query frame options are considered, while those of the reference frame are not.

In surjective matching, the matching is performed flexibly as there are no restrictions on the matching process. This allows surjective matching to deal with visually different frames well but also makes it susceptible to background distractions. For example, if the query frame contains some background distractors that look similar to the target object, they will erroneously get high foreground scores as the information about the target object in the reference frame will be allocated to them too. To prevent this, KMN~\cite{KMN} and BMVOS~\cite{BMVOS} introduce bijective matching mechanisms to capture the locality of a video. The key strategy of those methods involves limiting the number of chances of each reference frame pixel to be referenced by the query frame pixels. To this end, KMN uses Gaussian kernelling based on a reference-wise argmax operation, while BMVOS employs a reference-wise top K selection. By strictly restricting the conditions for feature matching, they can handle cases with background distractions.

Although these methods can relieve the limitations of surjective matching, they suffer from two limitations. First, as the existing bijective matching methods are based on discrete functions (an argmax or a top K operation) that obstruct stable network training, they are only applied at test time substituting surjective matching. This prevents a full utilization of surjective matching that has different advantages to bijective matching. Second, as they are very sensitive to hyper-parameters, they need to be tuned manually at test time to determine the optimal settings.

To overcome these limitations, we introduce an equalized matching mechanism. When connecting the query frame pixels and the reference frame pixels, we first equalize the influence of every reference frame pixel by applying a softmax operation to each reference frame pixel. Through this, confusing reference frame pixels (usually background distractors) that are referenced too often will lower their matching scores as all matching scores should divide up the same pie. This makes equalized matching robust against background distractors, which accords with the objective of bijective matching. While being robust, it does not suffer from the issues faced by existing bijective matching methods, as it is end-to-end learnable, can stand alone, and does not have any hyper-parameters. We visualize the comparison of surjective matching, bijective matching, and the proposed equalized matching in Figure~\ref{figure1}.

On public benchmark datasets, we thoroughly validate our proposed equalized matching mechanism. If used alone, it shows comparable performance to bijective matching methods that are carefully tuned for the optimal performance. If used jointly with surjective matching, it outperforms existing bijective matching methods by a significant margin, thanks to complementary properties of surjective matching and equalized matching (flexibility and reliability). By simply plugging an equalized matching branch to the baseline model, we achieve a comparable performance to state-of-the-art methods.

Our main contributions can be summarized as follows:
\begin{itemize}[leftmargin=0.2in]
	\item We introduce an equalized matching mechanism to overcome the limitations of existing bijective matching methods, while sustaining the same advantages. 
	
	\item The proposed equalized matching can easily be plugged into any existing networks, as well as used as an independent branch or jointly with surjective matching. 
	
    \item On public benchmark datasets, we achieve competitive performance through a simple and intuitive approach. 
\end{itemize}

\section{Related Work}
\noindent\textbf{Feature similarity matching.} Considering the target objects are not pre-defined, most existing VOS methods are built based on feature similarity matching. VideoMatch~\cite{VideoMatch} produces foreground and background similarity maps by comparing the embedded features of the initial frame and the query frame at the pixel level. Extending VideoMatch, FEELVOS~\cite{FEELVOS} and CFBI~\cite{CFBI} use the previous as well as the initial frame for the query frame prediction. To fully exploit all past frames for prediction, STM~\cite{STM} proposes a memory network-based architecture. As storing all past frames in external memory increases memory consumption and computational cost, GC~\cite{GC} stores an updatable key--value relation instead of storing key and value, while AFB-URR~\cite{AFB-URR} selectively stores useful features instead of storing all the extracted features. Similarly, RDE-VOS~\cite{RDE-VOS} proposes a recurrent dynamic embedding to build a memory bank with a constant size, and SWEM~\cite{SWEM} proposes a sequential weighted expectation-maximization network to reduce the redundancy of memory features. Considering the lack of details when only employing high-level feature matching, HMMN~\cite{HMMN} proposes a novel hierarchical matching mechanism to capture small objects as well. To relieve potential errors that can be caused by employing a pixel-level template, AOC~\cite{AOC} employs an adaptive proxy-level template, and TBD~\cite{TBD} employs both pixel-level and object-level templates simultaneously.

\vspace{1mm}
\noindent\textbf{Bijective matching.} As conventional surjective matching only considers the options for the query frame, it is naturally susceptible to background distractions. To better deal with them, bijective matching methods are proposed for a stabler information transfer. KMN~\cite{KMN} captures the bijectiveness of the feature matching using Gaussian kernelling based on a reference-wise argmax operation. After identifying the best-matching query frame pixel for a reference frame pixel, a Gaussian kernel is applied centered on the selected query frame pixel. Through this, the matching scores for the pixels distant from the kernel center can be reduced. Pursuing a similar objective, i.e., suppressing background distractions, BMVOS~\cite{BMVOS} uses a reference-wise top K selection. By limiting the number of reference frame pixels to be referenced by the query frame pixels, the foreground information can be prevented from being transferred to the background distractors. However, as the existing bijective matching methods are based on discrete functions that destabilize the network training, they are only applied at the testing stage replacing surjective matching. Therefore, they cannot be used with surjective matching that has different advantages to bijective matching. In addition, as they have hyper-parameters to be carefully tuned, test-time manual tuning is required. To overcome these limitations, we design a novel equalized matching mechanism.

\section{Approach}
\subsection{Problem Formulation}
Using the target object information in the reference frame, the goal of semi-supervised VOS is to predict the query frame segmentation mask. To transfer the information from the reference frame to the query frame, we use the feature similarity matching between the reference frame features and the query frame features. Let us denote the RGB image, embedded features, and predicted (or ground truth) segmentation mask at frame $i$ as $I^i\in[0,255]^{3\times H0\times W0}$, $X^i\in\mathbb{R}^{C\times H\times W}$, and $M^i\in[0,1]^{2\times H0\times W0}$, respectively. A downsampled version of $M^i$ is also prepared as $m^i\in[0,1]^{2\times H\times W}$. Each channel in $M$ and $m$ indicates background or foreground probability map. Given that frame $k$ is the reference frame and frame $i$ is the query frame to be processed, the objective can be written as inferring $M^i$ using $X^k$, $m^k$, and $X^i$. Based on the dense correspondence between $X^k$ and $X^i$, target object information $m^k$ is transferred to the query frame.

\subsection{Surjective Matching}
\begin{figure}[t]
	\centering
	\includegraphics[width=1\linewidth]{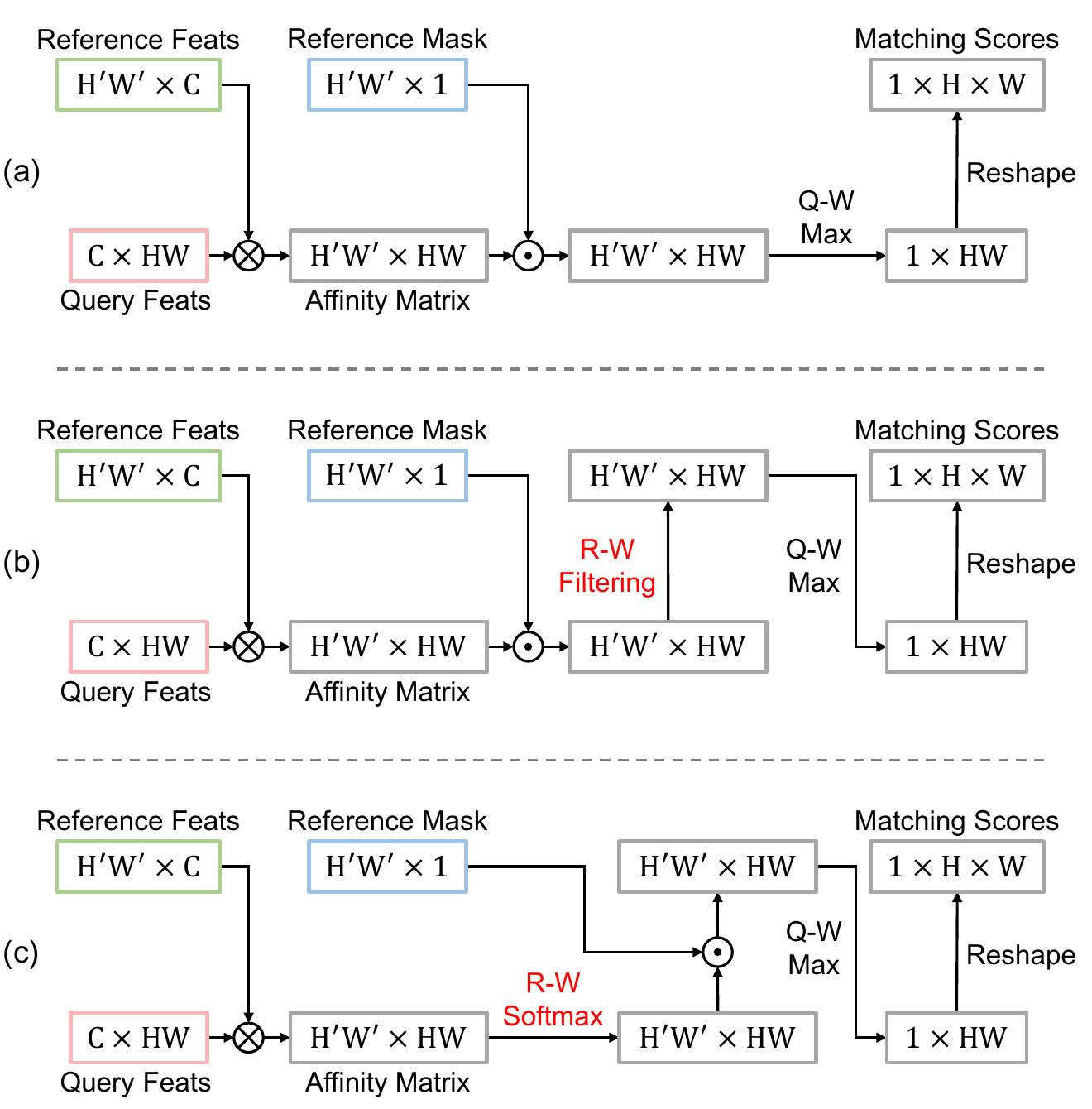}
	\caption{Comparison between (a) surjective matching, (b) bijective matching, and (c) equalized matching. Q-W and R-W indicate query-wise and reference-wise operations, respectively.}
	\label{figure2}
\end{figure}

A straightforward approach to transfer information of the reference frame to the query frame is to find the best matches from the query frame to the reference frame. Once the best matches are found, reference frame information, i.e., the reference frame segmentation mask, is transferred to the query frame for query frame mask prediction. To this end, we first compute the dense correspondence scores between the reference frame and query frame features. If $p$ and $q$ are the single pixel locations of the embedded features extracted from the reference frame $k$ and the query frame $i$, their feature similarity can be obtained using
\begin{align}
&sim\left(p, q\right) = \frac{\mathcal{N}\left(X^k_p\right) \cdot \mathcal{N}\left(X^i_q\right) + 1} {2}~,
\label{eq1}
\end{align}
where $\cdot$ denotes matrix inner product and $\mathcal{N}$ denotes channel L2 normalization. Note that $H$ and $W$ of the reference frame are notated as $H'$ and $W'$ for a better clarity. Instead of using a naive cosine similarity calculation, we apply linear normalization to force the scores to be in the same range as the mask scores. After calculating the similarities between every $p$ and $q$ in the reference and query frames, the affinity matrix $A\in[0,1]^{H'W'\times HW}$ can be obtained. To embody the target object information, the segmentation mask is multiplied to $A$ as
\begin{align}
&A_{BG} = A \odot M^k_0\nonumber\\
&A_{FG} = A \odot M^k_1~,
\label{eq2}
\end{align}
where $\odot$ indicates Hadamard product. After that, by applying query-wise maximum operation to $A_{BG}$ and $A_{FG}$, the matching score maps of the background and foreground can be obtained. By concatenating those score maps along channel, the final score map for frame $i$ can be defined as $S^i\in[0,1]^{2\times H\times W}$. The visualized pipeline of the surjective matching is also described in Figure~\ref{figure2}~(a).

As surjective matching only considers the query frame options, the matching can be performed flexibly. This enables surjective matching to transfer information effectively between visually different frames, i.e., temporally-distant frames, but makes it vulnerable to background distractors that need to be handled strictly.

\subsection{Bijective Matching}
Instead of providing absolute options to the query frame, the options of both the reference frame and the query frame should be considered to better deal with the background distractions. To this end, KMN~\cite{KMN} and BMVOS~\cite{BMVOS} introduce bijective matching methods that dynamically restrict the query frame options for stricter and more reliable information transfer. Before the query-wise maximum operation, a pre-defined filtering map, such as a Gaussian kernel or a top K mask, is multiplied to the class-embodied affinity matrix $B$, i.e., $A_{BG}$ or $A_{FG}$, as
\begin{eqnarray}
&B_p \leftarrow f(p) \odot B_p~,
\end{eqnarray}
where $f$ indicates the filter generation function that outputs the values from 0 to 1. As the query frame options are now partial rather than absolute, the matching scores can be implemented to follow the pre-defined design goals, e.g., excluding distant pixels from the best-matching pixel or discarding pixels except best-matching K pixels. The pipeline of bijective matching is also visualized in Figure~\ref{figure2}~(b).

Although bijective matching methods can reflect reference frame options in the feature similarity scores, they can only be adopted at the testing stage replacing surjective matching, as they are based on discrete functions that restrain stable network training. Therefore, they cannot be used jointly with surjective matching that has different advantages to bijective matching. Furthermore, test-time manual tuning is required for searching the optimal hyper-parameter settings.

\subsection{Equalized Matching}
To overcome these limitations of the existing bijective matching methods while maintaining the same objective, we design a novel matching method that is totally independent to surjective matching and still can capture bijectiveness during feature matching. Here, we introduce an equalized matching mechanism that is fully differentiable, and therefore, can stand alone as a bold branch or be used with surjective matching simultaneously. As in surjective matching and bijective matching, feature similarity between $p$ and $q$ is first computed as
\begin{eqnarray}
&sim\left(p, q\right) = X^k_p \cdot X^i_q~.
\end{eqnarray}
Based on the feature similarity, the affinity matrix $A$ can be calculated as before. Then, to impress the bijectiveness in the affinity matrix itself, a softmax operation is performed along the query dimension as
\begin{eqnarray}
&A_p \leftarrow Softmax(A_p)~.
\end{eqnarray}
Through this process, the sum of all query frame pixels' scores becomes 1 for each reference frame pixel, i.e., all reference frame pixels have the same contribution to the query frame prediction. Therefore, if a reference frame pixel is referenced a lot, its scores will be lowered as they should divide up the same pie. Considering that such confusing reference frame pixels may cause critical errors, strictly suppressing their matching scores is an effective strategy for minimizing visual distractions. After modulating the affinity matrix $A$, the remaining process is implemented just as in surjective matching, as described in Figure~\ref{figure2}~(c).

Unlike existing bijective matching methods, the proposed equalized matching mechanism does not contain discrete functions, so it can be stably learned during the network training stage as an independent branch. It can play the same role as bijective matching methods but does not suffer from the same problems as them. First, it can be flexibly plugged on top of surjective matching as well as used alone, enabling a full exploitation of surjective matching. Second, test-time manual tuning is not required as no hyper-parameters are needed.

\begin{figure}[t]
	\centering
	\includegraphics[width=1\linewidth]{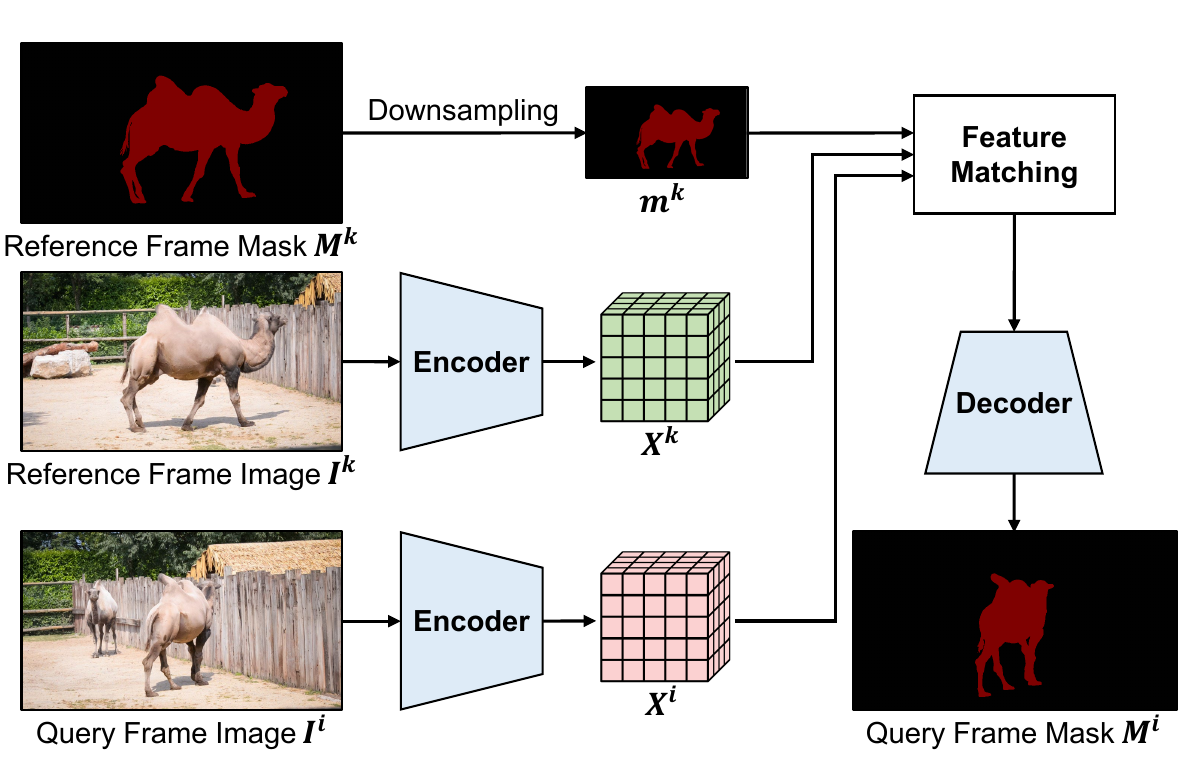}
	\caption{Architecture of our proposed method. The query frame is matched to the reference frame using the selected matching method. For simplicity and better understanding, mask propagation and skip connections are omitted in the illustration.}
	\label{figure3}
\end{figure}

\subsection{Network Architecture}
Our network is based on a simple encoder–decoder architecture as illustrated in Figure~\ref{figure3}. At the query frame, embedded features are first extracted from an image using an encoder. Then, those features are compared to the reference frame features via feature similarity matching. As reference frames, initial frame and previous adjacent frames are selected. For feature matching, we use surjective matching and equalized matching simultaneously, instead of transforming surjective matching to bijective matching at test time. This enables our network to capture flexibility as well as reliability through a visual matching process. Finally, by decoding the embedded features with the generated four score maps, query frame mask can be inferred.

\subsection{Implementation Details}
\noindent\textbf{Encoder.} We use DenseNet-121~\cite{densenet}, which is pre-trained on ImageNet~\cite{imagenet}, as our backbone network. It can be broadly divided into three blocks, which output 1/4-, 1/8-, and 1/16-scaled feature maps compared to the input image resolution. Only the 1/16-scaled feature maps are used for feature similarity matching.

\vspace{1mm}
\noindent\textbf{Decoder.} The decoder is designed identically to TBD~\cite{TBD}. It consists of convolutional layers that fuse and refine different features, and deconvolutional layers~\cite{deconv} that upscale the refined features. For fast decoding, every deconvolutional layer outputs only two channels to reduce the computational complexity. Before every deconvolutional layer, CBAM~\cite{CBAM} is added to reinforce feature representations.

\begin{table}[t]
	\centering 
	\caption{Quantitative evaluation on the DAVIS 2016 validation set. OL denotes online learning. (+S) denotes the use of static image datasets during the network training.}
	\vspace{2mm}
	\small
	\begin{tabular}{p{2.8cm}P{0.58cm}P{0.58cm}P{0.58cm}P{0.58cm}P{0.58cm}}
		\toprule
		Method &OL &fps &$\mathcal{G}_\mathcal{M}$ &$\mathcal{J}_\mathcal{M}$ &$\mathcal{F}_\mathcal{M}$\\
		\midrule
		STCNN~(+S)~\cite{STCNN} & &0.26 &83.8 &83.8 &83.8\\
		FEELVOS~(+S)~\cite{FEELVOS} &  &2.22 &81.7 &80.3 &83.1\\
		RANet~(+S)~\cite{RANet} & &30.3 &85.5 &85.5 &85.4\\
		DTN~(+S)~\cite{DTN} & &14.3 &83.6 &83.7 &83.5\\
		STM~(+S)~\cite{STM} & &6.25 &86.5 &84.8 &88.1\\
		DIPNet~(+S)~\cite{DIPNet} & &0.92 &86.1 &85.8 &86.4\\
		CFBI~(+S)~\cite{CFBI} & &5.56 &86.1 &85.3 &86.9\\
		GC~(+S)~\cite{GC} & &25.0 &86.6 &87.6 &85.7\\
		KMN~(+S)~\cite{KMN} & &8.33 &87.6 &87.1 &88.1\\
        STG-Net~(+S)~\cite{STG-Net} & &- &85.7 &85.4 &86.0\\
		RMNet~(+S)~\cite{RMNet} & &11.9 &81.5 &80.6 &82.3\\
        HMMN~(+S)~\cite{HMMN} & &10.0 &\textbf{89.4} &\textbf{88.2} &\textbf{90.6}\\
		\rowcolor{Gray}
		EMVOS~(+S) & &\textbf{49.8} &\underline{88.4} &\underline{87.9} &\underline{88.9}\\
		\hline
		RANet~\cite{RANet} & &30.3 &- &73.2 &-\\
		FRTM~\cite{FRTM} &\checkmark &21.9 &81.7 &- &-\\
		BMVOS~\cite{BMVOS} & &\underline{45.9} &\underline{82.2} &\underline{82.9} &\underline{81.4}\\
		\rowcolor{Gray}
		EMVOS & &\textbf{49.8} &\textbf{86.0} &\textbf{86.7} &\textbf{85.3}\\
		\hline
	\end{tabular}
	\label{Table:DAVIS16}
\end{table}

\subsection{Network Training}
\noindent\textbf{Pre-training on static images.} Following the existing VOS solutions~\cite{RANet, STM, GC, AFB-URR, HMMN}, we simulate videos by augmenting static images to obtain diverse network training samples. To this end, an image segmentation dataset COCO~\cite{COCO} is used. For each image sample, simulated video sequence is generated by augmenting the source image. The simulated video sequence has 5 frames, i.e., a source image and 4 sequentially-augmented images.

\vspace{1mm}
\noindent\textbf{Main training on videos.} After pre-training the network on the simulated samples, the network is trained on either the DAVIS 2017~\cite{DAVIS2017} training set or the YouTube-VOS 2018~\cite{YTVOS} training set depending on the target testing dataset. During this main training stage, we randomly sample 10 consecutive frames.

\vspace{1mm}
\noindent\textbf{Detailed settings.} In all training stages, video frames are randomly cropped to have 384$\times$384 resolution. For stable network training, balanced random cropping is adopted as in CFBI~\cite{CFBI} and TBD~\cite{TBD}. To provide challenging but natural training samples, the swap-and-attach augmentation is applied with a probability of 20\%, as in TBD. The learning rate is set to 1e-4 without learning rate decay, and Adam optimizer~\cite{adam} is used. Batch normalization layers~\cite{batchnorm} are also frozen following common protocol.

\begin{table}[t]
	\centering 
	\caption{Quantitative evaluation on the DAVIS 2017 validation set. OL denotes online learning. (+S) denotes the use of static image datasets during the network training.}
	\vspace{2mm}
	\small
	\begin{tabular}{p{2.8cm}P{0.58cm}P{0.58cm}P{0.58cm}P{0.58cm}P{0.58cm}}
		\toprule
		Method &OL &fps &$\mathcal{G}_\mathcal{M}$ &$\mathcal{J}_\mathcal{M}$ &$\mathcal{F}_\mathcal{M}$\\
		\midrule
		STCNN~(+S)~\cite{STCNN} & &0.26 &61.7 &58.7 &64.6\\
		FEELVOS~(+S)~\cite{FEELVOS} & &2.22 &69.1 &65.9 &72.3\\
		DMM-Net~(+S)~\cite{DMM-Net} & &- &70.7 &68.1 &73.3\\
		AGSS-VOS~(+S)~\cite{AGSS-VOS} & &10.0 &67.4 &64.9 &69.9\\
		RANet~(+S)~\cite{RANet} & &30.3 &65.7 &63.2 &68.2\\
		DTN~(+S)~\cite{DTN} & &14.3 &67.4 &64.2 &70.6\\
		STM~(+S)~\cite{STM} & &6.25 &71.6 &69.2 &74.0\\
		DIPNet~(+S)~\cite{DIPNet} & &0.92 &68.5 &65.3 &71.6\\
		LWL~(+S)~\cite{LWL} &\checkmark &14.0 &74.3 &72.2 &76.3\\
		CFBI~(+S)~\cite{CFBI} & &5.56 &74.9 &72.1 &77.7\\
		GC~(+S)~\cite{GC} & &25.0 &71.4 &69.3 &73.5\\
		KMN~(+S)~\cite{KMN} & &8.33 &76.0 &74.2 &77.8\\
		AFB-URR~(+S)~\cite{AFB-URR} & &4.00 &74.6 &73.0 &76.1\\
        STG-Net~(+S)~\cite{STG-Net} & &- &74.7 &71.5 &77.9\\
		RMNet~(+S)~\cite{RMNet} & &11.9 &75.0 &72.8 &77.2\\
		LCM~(+S)~\cite{LCM} & &8.47 &75.2 &73.1 &77.2\\
		SSTVOS~(+S)~\cite{SSTVOS} & &- &78.4 &75.4 &\underline{81.4}\\
        HMMN~(+S)~\cite{HMMN} & &10.0 &\textbf{80.4} &\textbf{77.7} &\textbf{83.1}\\
		\rowcolor{Gray}
		EMVOS~(+S) & &\textbf{49.8} &\underline{79.0} &\underline{76.9} &81.2\\
		\hline
		AGSS-VOS~\cite{AGSS-VOS} & &10.0 &66.6 &63.4 &69.8\\
		STM~\cite{STM} & &6.25 &43.0 &38.1 &47.9\\
		FRTM~\cite{FRTM} &\checkmark &21.9 &68.8 &66.4 &71.2\\
		TVOS~\cite{TVOS} & &37.0 &72.3 &69.9 &74.7\\
		JOINT~\cite{JOINT} &\checkmark &4.00 &\textbf{78.6} &\textbf{76.0} &\textbf{81.2}\\
		BMVOS~\cite{BMVOS} & &\underline{45.9} &72.7 &70.7 &74.7\\
		\rowcolor{Gray}
		EMVOS & &\textbf{49.8} &\underline{75.6} &\underline{73.8} &\underline{77.5}\\
		\hline
	\end{tabular}
	\label{Table:DAVIS17}
\end{table}

\section{Experiments}
The datasets and evaluation metrics used in this study are described in Section~\ref{setup}. Quantitative comparison to the state-of-the-art VOS solutions can be found in Section~\ref{quanti}. To validate our proposed approach, we analyze each proposed component thoroughly in Section~\ref{analysis}. Note that our method is abbreviated as EMVOS and all our experiments are implemented on a single GeForce RTX 2080 Ti GPU.

\subsection{Experimental Setup}
\label{setup}
\noindent\textbf{Datasets.} We use the DAVIS~\cite{DAVIS2016, DAVIS2017} and YouTube-VOS~\cite{YTVOS} datasets to validate our proposed approach. DAVIS 2016 and DAVIS 2017 respectively contain 50 and 120 video sequences having 24 fps. YouTube-VOS 2018 is the largest dataset for VOS, comprising 3,945 video sequences having 30 fps. All frames are annotated in the DAVIS datasets, while every five frames are annotated in the YouTube-VOS dataset.

\vspace{1mm}
\noindent\textbf{Evaluation metrics.} We use intersection-over-union (IoU) to evaluate the segmentation performance. $\mathcal{J}$ measure denotes the IoU for the entire region of an object, while $\mathcal{F}$ measure denotes the IoU for only the boundaries of an object. $\mathcal{G}$ measure, which is the average of $\mathcal{J}$ and $\mathcal{F}$, is used as the primary measure for evaluating VOS performance.

\begin{table}[t]
	\centering 
	\caption{Quantitative evaluation on the DAVIS 2017 test-dev set. OL denotes online learning. (+S) denotes the use of static image datasets during the network training.}
	\vspace{2mm}
	\small
	\begin{tabular}{p{2.8cm}P{0.58cm}P{0.58cm}P{0.58cm}P{0.58cm}P{0.58cm}}
		\toprule
		Method &OL &fps &$\mathcal{G}_\mathcal{M}$ &$\mathcal{J}_\mathcal{M}$ &$\mathcal{F}_\mathcal{M}$\\
		\midrule
        OSVOS-S~(+S)~\cite{OSVOS-S} &\checkmark &0.22 &57.5 &52.9 &62.1\\
		CNN-MRF~(+S)~\cite{CNN-MRF} &\checkmark &0.03 &67.5 &64.5 &\underline{70.5}\\
		DyeNet~(+S)~\cite{DyeNet} &\checkmark &0.43 &\underline{68.2} &\underline{65.8} &\underline{70.5}\\
		PReMVOS~(+S)~\cite{PReMVOS} &\checkmark &0.03 &\textbf{71.6} &\textbf{67.5} &\textbf{75.7}\\		
		FEELVOS~(+S)~\cite{FEELVOS} & &2.22 &54.4 &51.2 &57.5\\
		AGSS-VOS~(+S)~\cite{AGSS-VOS} & &10.0 &57.2 &54.8 &59.7\\
		RANet~(+S)~\cite{RANet} & &30.3 &55.3 &53.4 &57.2\\
        STG-Net~(+S)~\cite{STG-Net} & &- &63.1 &59.7 &66.5\\ 
		\rowcolor{Gray}
		EMVOS~(+S) & &\textbf{49.8} &67.5 &65.4 &69.6\\
		\hline
		AGSS-VOS~\cite{AGSS-VOS} & &10.0 &54.3 &51.5 &57.1\\
		TVOS~\cite{TVOS} & &37.0 &\underline{63.1} &58.8 &\textbf{67.4}\\
		BMVOS~\cite{BMVOS} & &\underline{45.9} &62.7 &\underline{60.7} &64.7\\
		\rowcolor{Gray}
		EMVOS & &\textbf{49.8} &\textbf{64.0} &\textbf{61.9} &\underline{66.1}\\
		\hline
	\end{tabular}
	\label{Table:DAVIS17t}
\end{table}

\subsection{Quantitative Results}
\label{quanti}
\noindent\textbf{DAVIS.} We compare our method to other state-of-the-art methods on the DAVIS~\cite{DAVIS2016, DAVIS2017} datasets in Table~\ref{Table:DAVIS16}, Table~\ref{Table:DAVIS17}, and Table~\ref{Table:DAVIS17t}. On all datasets, the inference speed is calculated on the DAVIS 2016 validation set, and 480p resolution is used. If static image datasets are used for network training, EMVOS achieves the second place on the DAVIS 2016 validation set and DAVIS 2017 validation set. On the DAVIS 2017 test-dev set, it is only outperformed by the online learning-based DyeNet~\cite{DyeNet} and PReMVOS~\cite{PReMVOS}, which are significantly slower than our method. If only the DAVIS training set is used for network training, EMVOS outperforms all other methods on the DAVIS 2016 validation set and DAVIS 2017 test-dev set. Considering its high inference speed of 49.8 fps, EMVOS shows the best speed--accuracy trade-off among all existing VOS methods on the DAVIS datasets.

\begin{table}[t]
	\centering 
	\caption{Quantitative evaluation on the YouTube-VOS 2018 validation set. (+S) denotes the use of static image datasets during the network training.}
	\vspace{2mm}
	\small
	\begin{tabular}{p{2.2cm}P{0.52cm}P{0.52cm}P{0.52cm}P{0.52cm}P{0.52cm}P{0.52cm}}
		\toprule
		Method &fps &$\mathcal{G}_\mathcal{M}$ &$\mathcal{J}_\mathcal{S}$ &$\mathcal{J}_\mathcal{U}$ &$\mathcal{F}_\mathcal{S}$ &$\mathcal{F}_\mathcal{U}$\\
		\midrule
		STM~(+S)~\cite{STM} &- &79.4 &79.7 &72.8 &84.2 &80.9\\
		SAT~(+S)~\cite{SAT} &39.0 &63.6 &67.1 &55.3 &70.2 &61.7\\
		LWL~(+S)~\cite{LWL} &- &81.5 &80.4 &76.4 &84.9 &84.4\\
        EGMN~(+S)~\cite{EGMN} &- &80.2 &80.7 &74.0 &85.1 &80.9\\
		CFBI~(+S)~\cite{CFBI} &- &81.4 &81.1 &75.3 &85.8 &83.4\\
		GC~(+S)~\cite{GC} &- &73.2 &72.6 &68.9 &75.6 &75.7\\
		KMN~(+S)~\cite{KMN} &- &81.4 &81.4 &75.3 &85.6 &83.3\\
		RMNet~(+S)~\cite{RMNet} &- &81.5 &82.1 &75.7 &85.7 &82.4\\
		LCM~(+S)~\cite{LCM} &- &82.0 &82.2 &75.7 &86.7 &83.4\\
		GIEL~(+S)~\cite{GIEL} &- &80.6 &80.7 &75.0 &85.0 &81.9\\
		SwiftNet~(+S)~\cite{SwiftNet} &- &77.8 &77.8 &72.3 &81.8 &79.5\\
		SSTVOS~(+S)~\cite{SSTVOS} &- &81.7 &81.2 &76.0 &- &-\\
		JOINT~(+S)~\cite{JOINT} &- &83.1 &81.5 &\textbf{78.7} &85.9 &\underline{86.5}\\
		HMMN~(+S)~\cite{HMMN} &- &82.6 &82.1 &76.8 &87.0 &84.6\\
		AOT-T~(+S)~\cite{AOT} &\underline{41.0} &80.2 &80.1 &74.0 &84.5 &82.2\\
		AOT-L~(+S)~\cite{AOT} &16.0 &\underline{83.8} &\underline{82.9} &77.7 &\textbf{87.9} &\underline{86.5}\\
		STCN~(+S)~\cite{STCN} &- &83.0 &81.9 &77.9 &86.5 &85.7\\
		RPCMVOS~(+S)~\cite{RPCMVOS} &- &\textbf{84.0} &\textbf{83.1} &\underline{78.5} &\underline{87.7} &\textbf{86.7}\\
		\rowcolor{Gray}
		EMVOS~(+S) &28.6 &79.7 &79.3 &74.3 &83.4 &81.7\\
		\hline
		RVOS~\cite{RVOS} &22.7 &56.8 &63.6 &45.5 &67.2 &51.0\\
        A-GAME~\cite{A-GAME} &- &66.1 &67.8 &60.8 &- &-\\
		AGSS-VOS~\cite{AGSS-VOS} &12.5 &71.3 &71.3 &65.5 &76.2 &73.1\\
		CapsuleVOS~\cite{CapsuleVOS} &13.5 &62.3 &67.3 &53.7 &68.1 &59.9\\
		STM~\cite{STM} &- &68.2 &- &- &- &-\\
		FRTM~\cite{FRTM} &- &72.1 &72.3 &65.9 &76.2 &74.1\\
		TVOS~\cite{TVOS} &37.0 &67.8 &67.1 &63.0 &69.4 &71.6\\
		STM-cycle~\cite{STM-cycle} &\textbf{43.0} &69.9 &71.7 &61.4 &75.8 &70.4\\
		BMVOS~\cite{BMVOS} &28.0 &\underline{73.9} &\underline{73.5} &\underline{68.5} &\underline{77.4} &\underline{76.0}\\
		\rowcolor{Gray}
		EMVOS &28.6 &\textbf{78.6} &\textbf{78.5} &\textbf{73.1} &\textbf{82.5} &\textbf{80.3}\\
		\hline
	\end{tabular}
	\label{Table:YTVOS}
\end{table}

\vspace{1mm}
\noindent\textbf{YouTube-VOS.} The comparison of our method and other state-of-the-art methods on YouTube-VOS 2018~\cite{YTVOS} validation set is presented in Table~\ref{Table:YTVOS}. If external training data is adopted, heavy networks with a growing memory pool, such as RPCMVOS~\cite{RPCMVOS}, AOT~\cite{AOT}, JOINT~\cite{JOINT}, and STCN~\cite{STCN}, show notable performance. EMVOS achieves a comparable inference speed and segmentation accuracy to a lite version of AOT, with 28.6 fps and a $\mathcal{G}$ score of 79.7\%. Under the constraint of not using the external datasets, EMVOS outperforms all previous methods by a large margin, with a $\mathcal{G}$ score of 78.6\%. This supports the exceptional generalization ability of EMVOS; performance degradation is only 1.1\% while that of STM~\cite{STM} is 11.2\%.

\begin{table}[h!]
	\centering 
	\caption{Comparison of various model versions with various feature matching methods. SM, BM, and EM indicate surjective matching, bijective matching, and equalized matching, respectively. K1 and K2 indicate the hyper-parameters (kernel standard deviation or number of pixels to be filtered) for bijective matching with the initial and previous adjacent frames. The models are tested on the DAVIS 2017 validation set.}
	\vspace{2mm}
	\small
	\begin{tabular}{c|c|P{0.8cm}|P{0.8cm}|P{0.8cm}}
		\toprule
		Version &Matching &K1 &K2 &$\mathcal{G}_\mathcal{M}$\\
		\midrule
		Baseline &SM &- &- &75.7\\
        \midrule
        \Romannum{1} & SM $\rightarrow$ BM (Kernel) &16 &32 &76.5\\
        \Romannum{1} & SM $\rightarrow$ BM (Kernel) &32 &32 &77.5\\
        \Romannum{1} & SM $\rightarrow$ BM (Kernel) &64 &32 &76.5\\
        \Romannum{1} & SM $\rightarrow$ BM (Kernel) &32 &16 &77.3\\
        \Romannum{1} & SM $\rightarrow$ BM (Kernel) &32 &64 &77.3\\
        \midrule
        \Romannum{6} & SM $\rightarrow$ BM (Top K) &64 &16 &74.4\\
        \Romannum{6} & SM $\rightarrow$ BM (Top K) &128 &16 &76.2\\
        \Romannum{6} & SM $\rightarrow$ BM (Top K) &256 &16 &75.9\\
        \Romannum{6} & SM $\rightarrow$ BM (Top K) &128 &8 &74.5\\
        \Romannum{6} & SM $\rightarrow$ BM (Top K) &128 &32 &75.9\\
        \midrule
        \Romannum{11} & EM &- &- &77.9\\
        \Romannum{12} & SM + EM &- &- &79.0\\
		\bottomrule
	\end{tabular}
	\label{Table:ablation}
\end{table}

\begin{figure*}[t]
	\centering
	\includegraphics[width=1\linewidth]{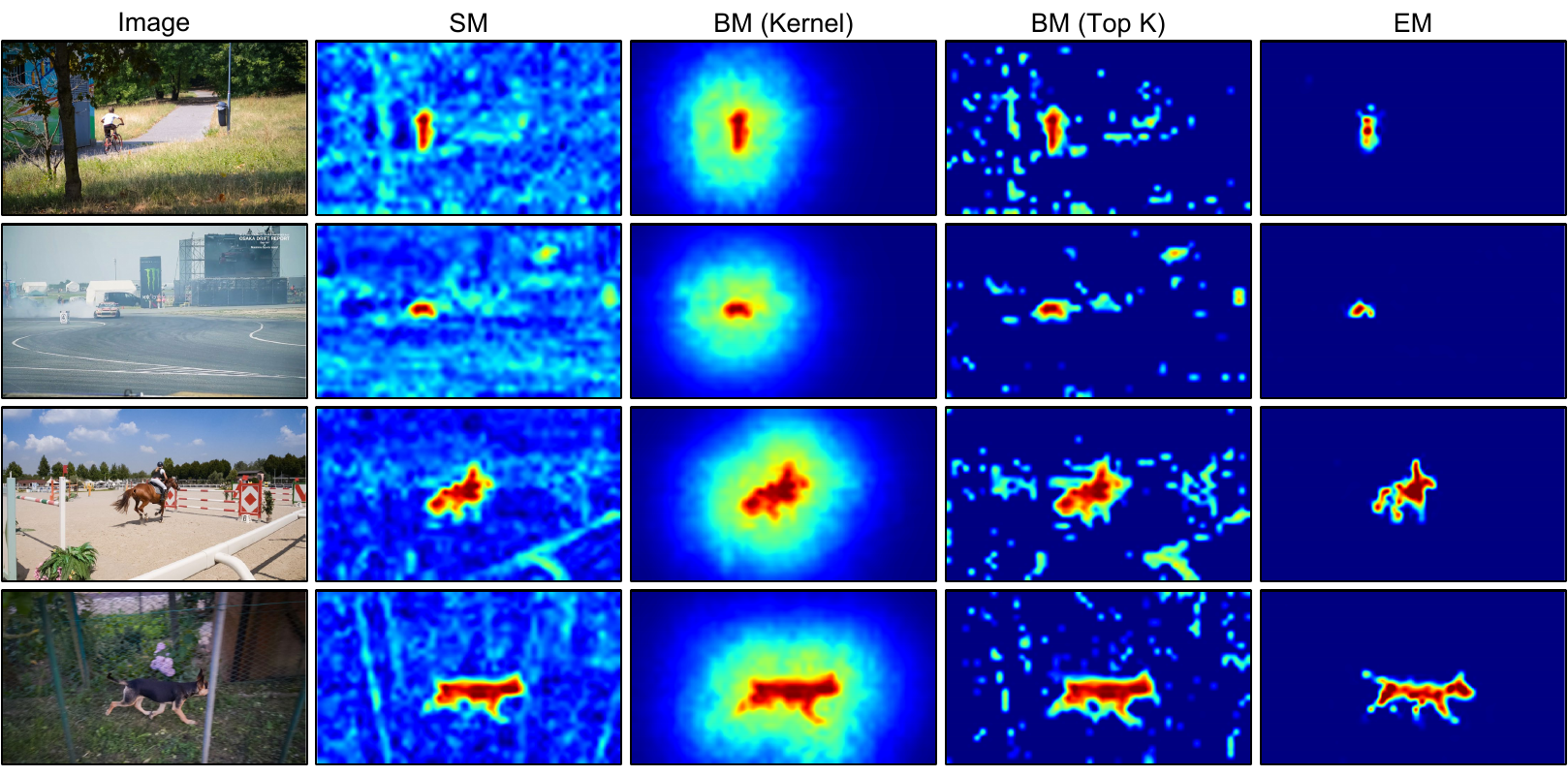}
	\caption{Comparison of the feature matching results obtained by various matching methods. SM, BM, and EM indicate surjective matching, bijective matching, and equalized matching, respectively. For a clear visualization, the values in each score map are linearly normalized to be in the same range.}
	\label{figure4}
\end{figure*}

\begin{figure*}[t]
	\centering
	\includegraphics[width=1\linewidth]{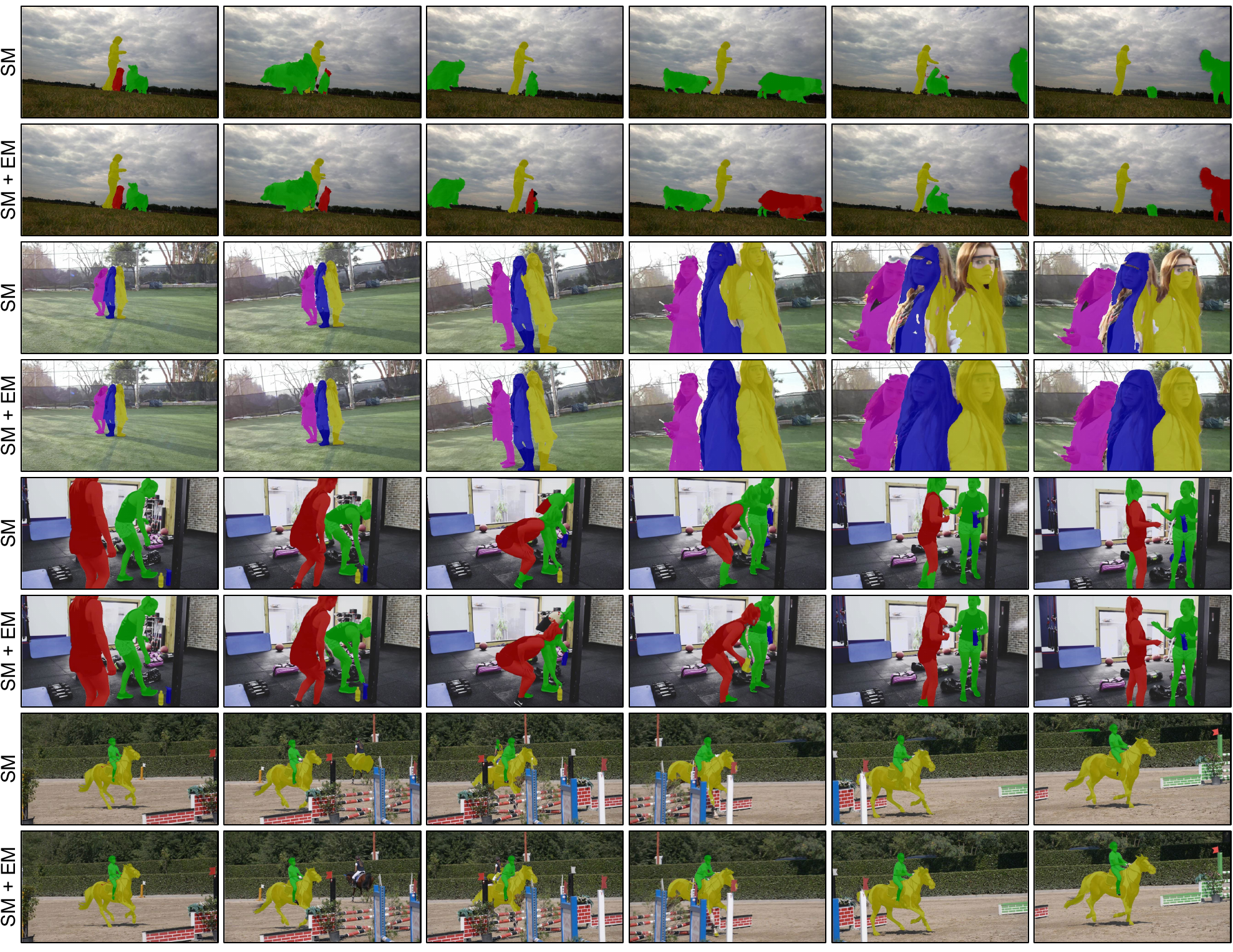}
	\caption{Qualitative segmentation mask comparison of the models using various matching methods. SM and EM indicate surjective matching and equalized matching, respectively.}
	\label{figure5}
\end{figure*}

\subsection{Analysis}
\label{analysis}
\noindent\textbf{Qualitative matching comparison.} To validate the effectiveness of our proposed equalized matching mechanism, we qualitatively compare the matching score maps of various matching methods in Figure~\ref{figure4}. As a naive surjective matching only considers the query frame options, all query frame pixels can make their best choices. Therefore, background distractions as well as the foreground object have high matching scores, which prevents a clear distinction between foreground and background. For a better separation of foreground and background, bijective matching methods apply some filtering strategies to surjective matching at test time. The kernel-based bijective matching method~\cite{KMN} can restrict the search area by applying a Gaussian kernel, but the score maps are quite coarse and blurred because of the nature of the kernelling. The top K-based bijective matching method~\cite{BMVOS} can suppress the background scores to some extent, but still suffers from noisy background distractors. By contrast, our proposed equalized matching method can clearly distinguish foreground objects from the background while maintaining fine details. Specifically, in the third and the fourth sequences, the thin parts of the horse and dog (tails and legs) are not well captured in surjective matching, kernel-based bijective matching, and top K-based bijective matching. However, in equalized matching, they can be well captured as information transfer is performed in a more strict and reliable manner.

\vspace{1mm}
\noindent\textbf{Quantitative mask comparison.} We also compare various matching methods with respect to their final segmentation performance. In Table~\ref{Table:ablation}, various model versions with various matching methods are quantitatively compared. The baseline model is a model employing naive surjective matching. It achieves a $\mathcal{G}$ score of 75.7\%. By adopting Gaussian kernelling or a top K selection to the baseline model at test time, further improvements can be achieved. The kernel-based bijective matching can bring up to 1.8\% performance improvement, and the top K-based bijective matching can boost up to 0.5\% on $\mathcal{G}$ metric. However, as can be seen from the table, bijective matching methods are quite sensitive to the hyper-parameter values. Consequently, they require an extra process for searching optimal hyper-parameters via test-time manual tuning, but this makes the solution less elegant and efficient. The proposed equalized matching method achieves a $\mathcal{G}$ score of 77.9\%, which surpasses the optimal performance of the bijective matching methods without the need for manual tuning. As well as being able to stand alone, it can also be used on top of surjective matching. If surjective matching and equalized matching are used together, the performance exceeds all other matching methods by a large margin, demonstrating the effectiveness of their joint use.

\vspace{1mm}
\noindent\textbf{Qualitative mask comparison.} In Figure~\ref{figure5}, output segmentation masks of model versions with various matching methods are qualitatively compared. The baseline model only uses surjective matching, while our final model leverages both surjective matching and equalized matching. In the first and the third sequences, the baseline model misses the red object because red and green objects look similar. As there is no limitation to the information-referencing process in surjective matching, the red object can get a high "green object score", which leads to the red object getting missed. In the second sequence, where the three target objects are visually similar, the baseline model fails to accurately segment the objects, especially for boundaries of the objects. This is because surjective matching is susceptible to confusing information, so the boundary regions that have relatively unclear features are not handled well. In the fourth sequence, wrong detection of the background distractors is also observed, as they have similar appearance to the target objects. Only employing surjective matching is proven to make the network vulnerable to the visual distractions. By contrast, our final model can effectively deal with all these sequences, thanks to its strong ability to distinguish foreground objects from the background via reliable and stable information propagation.

\section{Conclusion}
To improve the conventional surjective matching mechanism for VOS, bijective matching methods have been proposed, but these have certain limitations as well. To overcome them while maintaining the same objective, we introduce an equalized matching mechanism. The effectiveness of the proposed equalized matching mechanism is thoroughly validated via extensive experiments. By simply adopting surjective matching and equalized matching with a naive encoder--decoder architecture, we achieve the best speed--accuracy trade-off on public benchmark datasets. We believe our approach will be widely used in future research.


{\small
\bibliographystyle{ieee_fullname}
\bibliography{egbib}
}

\end{document}